\documentclass[conference]{IEEEtran}

\usepackage{import}
\usepackage{hyperref}

\usepackage{cite} 

\usepackage[pdftex]{graphicx}
\graphicspath{{./images/}}
\DeclareGraphicsExtensions{.png} 

\usepackage{amsmath}
\usepackage{amssymb}
\usepackage{mathtools}

\DeclareMathOperator*{\argmax}{arg\,max}
\newcommand{\thickhat}[1]{\mathbf{\hat{\text{$#1$}}}}

\usepackage{subcaption}

\usepackage[table]{xcolor}

\usepackage{algorithm}
\usepackage{algpseudocode}

\makeatletter
\AfterEndEnvironment{algorithm}{\let\@algcomment\relax}
\AtEndEnvironment{algorithm}{\kern2pt\hrule\relax\vskip3pt\@algcomment}
\let\@algcomment\relax
\newcommand\algcomment[1]{\def\@algcomment{\footnotesize#1}}

\renewcommand\fs@ruled{\def\@fs@cfont{\bfseries}\let\@fs@capt\floatc@ruled
  \def\@fs@pre{\hrule height.8pt depth0pt \kern2pt}%
  \def\@fs@post{}%
  \def\@fs@mid{\kern2pt\hrule\kern2pt}%
  \let\@fs@iftopcapt\iftrue}
\makeatother

\title{Directed Evolution of Proteins via Bayesian Optimization in Embedding Space}

\author{
\IEEEauthorblockN{1\textsuperscript{st} Matouš Soldát}
\IEEEauthorblockA{
\textit{Department of Computer Science}\\
\textit{FEE, Czech Technical University in Prague}\\
Prague, Czech Republic\\
matous.soldat@uochb.cas.cz\\
}
\and
\IEEEauthorblockN{2\textsuperscript{nd} Jiří Kléma}
\IEEEauthorblockA{
\textit{Department of Computer Science}\\
\textit{FEE, Czech Technical University in Prague}\\
Prague, Czech Republic\\
klema@fel.cvut.cz\\
}
}

\date{July 2024}

\newcommand{\DOI}{10.1109/BIBM62325.2024.10822356}
\usepackage{fancyhdr}

\fancypagestyle{firstpage}{

\cfoot{\fontsize{9pt}{11pt}\selectfont $\copyright$ \the\year~IEEE. Personal use of this material is permitted. Permission from IEEE must be obtained for all other uses, in any current or future media, including reprinting/republishing this material for advertising or promotional purposes, creating new collective works, for resale or redistribution to servers or lists, or reuse of any copyrighted component of this work in other works.}
}

\begin{document}
\bstctlcite{IEEEexample:BSTcontrol}

\maketitle
\thispagestyle{firstpage}

\begin{abstract}
Directed evolution is an iterative laboratory process of designing proteins with improved function by iteratively synthesizing new protein variants and evaluating their desired property with expensive and time-consuming biochemical screening. Machine learning methods can help select informative or promising variants for screening to increase their quality and reduce the amount of necessary screening. In this paper, we present a novel method for machine-learning-assisted directed evolution of proteins which combines Bayesian optimization with informative representation of protein variants extracted from a pre-trained protein language model. We demonstrate that the new representation based on the sequence embeddings significantly improves the performance of Bayesian optimization yielding better results with the same number of conducted screening in total. At the same time, our method outperforms the state-of-the-art machine-learning-assisted directed evolution methods with regression objective.
\end{abstract}


\section{Introduction}
Protein engineering (PE) is the process of designing proteins with desired properties, such as improved stability, catalytic function, or specific binding affinity \cite{ali2016pe}. PE can be leveraged in industrial applications, environmental applications, medicine, nanobiotechnology, and other fields \cite{ali2016pe}. Because the functional properties of proteins are determined by their sequence of amino acids \cite{yang2019}, the task of PE translates to finding a sequence of amino acids with the desired properties/function.
However there is an infinite number of possible protein sequences and non-functional proteins dominate the sequence space \cite{yang2019}, which makes PE a challenging task.
One of the most widespread approaches to this issue is Directed Evolution (DE) \cite{wang2021}.
\par
DE is an iterative laboratory process of creating new biomolecules of desired properties, which mimics Darwinian evolution in a controlled environment \cite{wang2021}.
DE circumvents the problem of the vast protein-sequence space filled with non-functional sequences by iteratively mutating an existing protein (often called the wild-type variant) to improve its function \cite{wang2021}.
A DE iteration consists of two main steps: mutagenesis, in which parent molecule(s) are mutated and/or recombined to create a library of variants, and screening/selection, where high-quality variants are identified to form a new generation of parents with improved properties \cite{wang2021}. The quality of a given protein variant in terms of the desired property is reported as a numerical value termed fitness.
\par
The wet lab experiments associated with synthetization and screening of the mutated protein variants are expensive and time-consuming \cite{yang2019}. Because of this, the screening process is a common bottleneck of all DE methods. This motivates the employment of machine learning methods to minimize the amount of conducted screening while maximizing the highest obtained fitness.
Instead of discarding the low-fitness variants as in traditional DE, methods of machine-learning-assisted directed evolution (MLDE) incorporate information about all screened variants into a model which predicts a protein's fitness based on its sequence \cite{yang2019}. The model is then used to intelligently select new variants for screening which maximize the predicted fitness and/or minimize uncertainty in the model \cite{yang2019}.
\par
In this work, we propose a novel MLDE method, Bayesian Optimization in Embedding Space (BOES), which combines Bayesian optimization (BO) \cite{shahriari2015bo} with informative embedding of protein sequences extracted by a pre-trained protein language model (PPLM). BOES exploits a PPLM to extract informative embeddings of all variants in the sequence space and the BO procedure is conducted in the obtained embedding space. In each iteration, a Gaussian process (GP) model is fitted to the already screened variants and the next variant for screening is chosen by maximization of expected improvement (EI). To the best of our knowledge, this paper is the first work that successfully combines BO with a PPLM-extracted embedding. In the following text, we describe the new combination and demonstrate its applicability to efficient protein engineering.

\section{Related Work}
A wide variety of models have been applied to MLDE including simple linear regression models, decision trees/forests, kernel methods, Gaussian Process models, and deep learning \cite{yang2019}.
Existing MLDE methods typically employ regression methods which require an additional exploitation stage since they do not prioritize high-fitness variants during the training \cite{wu2019, wittmann2021ftmlde, qiu2021clade, qiu2022clade2, qin2023}. In contrast, BO corresponds to the objective of MLDE much more closely. As an optimization method, BO aims to maximize the fitness function in each iteration and no additional exploitation stage is required. Furthermore, BO is very data efficient, making it an ideal choice in problems where the evaluation of data points is costly and the objective function is multimodal \cite{shahriari2015bo}. Both of these properties are key difficulties in exploring protein fitness landscapes \cite{wu2016gb1, simoncini2018fitness}.
\par
BO guides the exploration-exploitation trade-off based on the selected acquisition function. 
Two notable acquisition functions are widely used in PE applications. The upper confidence bound (UCB) acquisition function selects the data point with the largest upper confidence bound for evaluation, prioritizing data points that are predicted to be both optimized and uncertain \cite{romero2013navigating}. The relative importance of the prediction and uncertainty can be manually controlled with a weighting parameter \cite{hie2020leveraging}. The second notable acquisition function, expected improvement (EI), selects the data point where the expectation over the possible values of the objective function is predicted to have the largest improvement over the current best observation \cite{shahriari2015bo}. Similarly to UCB, this approach also strikes a balance between prioritizing data points predicted to be optimized and unexplored data points where the prediction is uncertain. Both methods have been shown to be efficient in the number of function evaluations required to find the global optimum of multi-modal black-box objective functions \cite{srinivas2009gaussian, bull2011convergence}.
\par
UCB has been used in GP regression with a structure-based metric of similarity to provide a probabilistic description of the landscapes for various properties of proteins and to design a cytochrome P450
variant that is more than 5\,°C more thermostable than P450 variants previously optimized by different methods and 14\,°C more stable than the most stable parent from which it was made \cite{romero2013navigating}. In \cite{bedbrook2017machine}, GP classification and regression models were trained with UCB on expression and localization data from 218 channelrhodopsin \cite{deisseroth2017form} variants. Structural similarity obtained by aligning residue-residue contact maps of each variant and counting the number of identical contacts were used as a metric of sequence similarity. In addition to GP regression with UCB criterion, in \cite{greenhalgh2021machine}, the method first samples 20 variants from the sequence space that maximize the Gaussian mutual information. The sampled variants are used to fit the GP before the first iteration of sequential optimization. Lastly, in \cite{hie2020leveraging}, a GP trained with UCB is compared with other methods that model uncertainty differently or do not model uncertainty at all. \cite{hie2020leveraging} highlights GP-based methods as particularly useful and shows a consistently strong performance of the GP model. 
\par
A GP model with the EI acquisition function has been shown to outperform traditional DE methods in an \textit{in silico} experiment \cite{frisby2020fold}. The proposed method selects variants for evaluation in batches of 19 and uses the squared exponential kernel with Euclidean distances computed from one-hot encoding of the variants at mutated positions.
The recent optimization framework for protein DE, termed ODBO \cite{cheng2022odbo}, combines GP and EI acquisition function with a novel low-dimensional, function-value-based protein encoding strategy and prescreening outlier detection. A protein variant is represented by a feature vector, where each amino acid from the sequence is replaced by the mean or maximum value of the fitness measurements of all variants with the amino acid at that position. Then, in each iteration, the vector representations are inputted into the prescreening via \textit{Extreme Gradient Boosting Outlier Detection} (XGBOD) \cite{zhao2018xgbod} which filters out potential low fitness samples before the BO step. \cite{cheng2022odbo} argues that the novel representation creates a smoother local variable for regression while the prescreening aims to perform more efficient acquisitions in each iteration.
\par
Different protein sequence representations have been applied in BO-based MLDE methods and the advantage of informative representation has been previously demonstrated \cite{cheng2022odbo}.
The representations learned by PPLMs are known to carry useful information about the function of the variants \cite{rives2021esm1b, vig2020bertology, rao2020transformer, elnaggar2021prottrans}, which enables the definition of a sensible metric of distance between variants.
Furthermore, a key advantage of the PPLM-extracted embedding space over different informative input spaces is that no variants need to be screened for the construction of the input space, saving screening costs.
However, PPLM-extracted embeddings have been previously thought to be incompatible with a GP model and BO because of their high dimensionality \cite{yang2024active}. In this work, we show that if we limit the number of hyperparameters of the GP model by reducing the number of effective dimensions, BO can be employed in the embedding space to great effect.

\section{Problem Formulation}
The task of MLDE is formulated as black-box optimization with expensive objective function evaluation.
The objective can be formalized as finding the word $\thickhat{\boldsymbol x}$ from the set of all words $\mathcal{X}$ over an alphabet consisting of the twenty common amino acids \cite{lopez2020aa} which maximizes the objective function $f:\mathcal{X}\xrightarrow{}\mathbb{R}$,
\begin{equation}\label{eq:problem}
    \thickhat{\boldsymbol x} = \argmax_{\boldsymbol x \in \mathcal{X}} f(\boldsymbol x)
\end{equation}
The objective function $f$ represents the costly screening experiments. For a variant specified by word $\boldsymbol x \in \mathcal{X}$, $f$ returns the fitness of the variant.
In full generality, the set $\mathcal{X}$ is infinite. In MLDE literature, the problem is often simplified by considering only substitutions of the wild-type protein, limiting the set to $l^{20}$ variants, where $l$ is the protein's length. The problem is usually simplified further, by limiting the number of mutation positions $n$ to very few positions selected by an informed oracle as largely influential to the protein's function, resulting in $n^{20}$ variants. This is also the case in our in-silico experiments, where we use two datasets, each with $n = 4$ pre-selected mutation positions.

\section{Proposed Method}
BOES employs BO with a GP model to select variants for screening in an MLDE procedure by maximizing the expected improvement (EI). BO with EI objective function is ideal for MLDE application because it corresponds perfectly to the objective of MLDE in each iteration. That is, each new variant for screening is chosen to maximize the expectation of improvement in the best-so-far screened fitness. This ensures the optimal use of resources after each iteration and eliminates the need for a predefined screening budget, which is necessary when model regression methods are employed.
\par
Before running the BO procedure, BOES uses a PPLM to extract informative sequence embeddings of all variants. The GP model is provided with an embedding function $g:\mathcal{X}\xrightarrow{}\mathbb{E}$ and
models the fitness landscape
in the $m$-dimensional sequence embedding space $\mathbb{E}\coloneq\mathbb{R}^m$, where variants with similar embeddings are expected to have similar properties.
\par
The BOES algorithm is described in Alg. \ref{alg:boes}. The MLDE procedure starts with only the wild-type protein in the set of observations $D_1=\{(\boldsymbol x_{wt}, y_{wt})\}$. In each iteration of BO, the GP is fitted to the current set of observations (already screened variants), the EI acquisition function is evaluated at each data point (each variant) and the variant with maximal EI is selected, screened, and added to the set of observations with the observed fitness value.
\begin{algorithm}
\caption{BOES} \label{alg:boes}
\textbf{Input:} All variants $\mathcal X$\\
\textbf{Output:} Best screened variant $(\boldsymbol x, y)$
\begin{algorithmic}[1]
\State Initialize dataset $\mathcal D_1 \gets \{(\boldsymbol x_{wt}, f(\boldsymbol x_{wt}))\}$ with the wild-type protein
\State Fit the model $GP_1$ given $\mathcal D_1$
\For{$n = 1, 2, \dots, k$}
    \State Select new variant for screening by optimizing EI \[\boldsymbol x_{n + 1} \gets \argmax_{\boldsymbol x \in \mathcal{X}} EI(\boldsymbol x; GP_n)\]
    \State Screen it $\mathcal D_{n+1} \gets\mathcal D_n \cup \{(\boldsymbol x_{n + 1}, f(\boldsymbol x_{n+1}))\}$
    \State Fit the model $GP_{n+1}$ given $\mathcal D_{n+1}$
\EndFor

\State \Return best variant $(\boldsymbol x, y) \gets \argmax_{(\boldsymbol x, y) \in\mathcal D_{k+1}} y$

\end{algorithmic}
\algcomment{
\begin{tabular}{rl}
    $EI$ &\dots\hspace{2mm} Expected Improvement acquisition function.\\
    $GP_n$ &\dots\hspace{2mm} Gaussian process model fitted to dataset $D_n$.\\
    $f:\mathcal{X}\xrightarrow{}\mathbb{R}$ &\dots\hspace{2mm} Screening, assigns fitness to a variant.\\
\end{tabular}
}
\end{algorithm}

\section{Implementation}
The ESM-1b model \cite{rives2021esm1b} was chosen as the embedding extractor for its ability to produce informative embeddings \cite{rives2021esm1b} and the widespread use of the ESM family of PPLMs in MLDE-related literature \cite{hie2022evovelocity, qin2023, wittmann2021ftmlde}.
A plethora of other PPLMs exist \cite{lin2022esm2, elnaggar2021prottrans, hesslow2022rita, ferruz2022protgpt2}, which could be applied to BOES in the future. The goal of this work is to demonstrate that the combination of PPLMs and BO is a feasible and promising direction for MLDE. 
\par
A fundamental problem of employing BO in the embedding space of a PPLM, and the probable reason why this approach has not been successfully employed before, is that BO struggles with high dimensional input spaces \cite{shahriari2015bo, cheng2022odbo}. This is problematic because PPLM embeddings tend to have a size in orders of $10^2$ to $10^3$, depending on the architecture of the language model. This means that we are trying to run BO in an input space with potentially thousands of dimensions.
To solve this issue, BOES defines the GP model with a custom implementation of the Matérn 3/2 kernel $k:\mathbb{E}\times\mathbb{E}\xrightarrow{}\mathbb{R}$ and Euclidean distance $d:\mathbb{E}\times\mathbb{E}\xrightarrow{}\mathbb{R}$,
\begin{equation}\label{eq:kernel}
    k(\boldsymbol e, \boldsymbol e^\prime) = \exp(-\sqrt{3}d(\boldsymbol e, \boldsymbol e^\prime))(1+\sqrt{3}d(\boldsymbol e, \boldsymbol e^\prime))
\end{equation}
\begin{equation}\label{eq:dist}
    d(\boldsymbol e, \boldsymbol e^\prime) = \sqrt{(\boldsymbol e - \boldsymbol e^\prime)^T(\mathbf{I}\cdot\theta^2)(\boldsymbol e - \boldsymbol e^\prime)}
\end{equation}
to limit the effective number of dimensions to one, so that the surrogate model only fits one length scale hyperparameter $\theta$ instead of fitting an individual length scale for each dimension of the embedding $\boldsymbol e = g(\boldsymbol x) \in \mathbb{E}$ extracted from sequence $\boldsymbol x$.
\par
For the prior distribution of the singular length scale, a normal distribution with a mean of zero and a standard deviation $\sigma$ of $\frac{\sqrt{1280}}{3}$, truncated (and normalized) to the interval $[0;\infty)$, was used. $\sigma$ was chosen such that the diagonal across the high-dimensional embedding space corresponds approximately to $3\sigma$. Since the embedding space of the used model, ESM-1b, has 1280 dimensions and the absolute values of the elements in the protein embeddings rarely exceed 1 (0.3~\% of the elements from all GB1 embeddings have absolute values higher than 1), the size of the diagonal is roughly $\sqrt{1280}$.
\par
The GP model is defined with zero prior mean function $\mu_0:\mathcal{X}\xrightarrow{}0$. Zero variance $\sigma^2$ is used for noise, effectively removing noise from the model, because the experiments are conducted on a noiseless dataset.
The BO procedure is implemented with the BOSS.jl package \cite{boss}. The model is fitted with maximum likelihood estimation by the NEWUOA algorithm \cite{powell2006newuoa} with 20 starts in a multi-start setting and lower bound on the trust region radius $\rho_{end} = 10^{-4}$. The zero noise variance $\sigma^2$ is replaced with a very small positive value to ensure numerical stability of the model. To avoid wasting the screening budget on already screened variants, the value of the acquisition function computed for each already screened variant is replaced by zero before the next variant for screening is chosen. This ensures that the screened variants cannot be chosen again unless the acquisition function value of all variants in the sequence space is also zero, which is practically impossible.
\par
Code for the implemented MLDE procedures and DE simulation baselines, as well as the used datasets, are available at \url{https://github.com/soldatmat/PELLM}. The MLDE procedures were implemented in a unified modular framework for in silico DE, which is made available separately as the DESilico.jl package \cite{desilico}.

\section{Results}
This section presents the experimental settings used for evaluation of the proposed method including datasets, means of evaluation, baseline DE simulations, and comparison to other implemented methods and SOTA MLDE methods.
\subsection{Data}
Experiments were carried out in silico on two datasets. Each dataset maps the fitness landscape of a different wild-type protein. The datasets consist of variant-fitness pairs of nearly all possible variants of the wild-type protein mutated at 4 positions. The fitness of each unmeasured variant is assumed to be zero in all conducted experiments since the unmeasured variants are considered meaningless to biologists \cite{wittmann2021ftmlde, qin2023}. The mutation positions were selected as largely influential to the structure and function of the protein.
\par
\textbf{GB1 dataset} \cite{wu2016gb1} is a dataset of variants of the protein G domain B1, mutated at four positions with non-linear epistasis (V39, D40, G41, V54). Fitness values of GB1 variants represent the binding ability to the antibody IgG-Fc and range from 0.0 to 8.76. Fitness value of 1.0 corresponds to the binding ability of the wild-type protein.
\par
\textbf{PhoQ dataset} \cite{podgornaia2015phoq} consists of variants of protein kinase PhoQ obtained by mutating the wild-type sequence at four positions critical to the function of the protein (A284, V285, S288, T289). The fitness values refer to the phosphatase or kinase activity of different PhoQ variants and range from 0.0 to 133.59. Fitness of the wild-type protein kinase PhoQ is 3.29.

\subsection{Performance on the Wild-type Protein}
To evaluate the performance of BOES, we compared its performance to SOTA regression-based MLDE methods, which all minimize the prediction error of the model during the DE procedure. The comparison in Table \ref{tbl:afpde_comparison} includes results of SOTA MLDE methods reported in \cite{qin2023}. A concise description of the included methods is adapted from \cite{qin2023}:
\begin{itemize}
    \item MLDE \cite{wu2019} trains an ensemble of shallow neural networks as fitness predictors on randomly sampled variants.
    \item ftMLDE, focused training MLDE \cite{wittmann2021ftmlde}, is a strategy for running MLDE with training sets designed to avoid holes. The comparison includes ftMLDE with two sampling strategies, EVmutation \cite{hopf2017mutation} and MSA-transformer \cite{rao2021msa}.
    \item CLADE \cite{qiu2021clade} trains a fitness predictor with high-fitness mutants obtained through a hierarchical clustering sampling method.
    \item CLADE 2.0 \cite{qiu2022clade2} selects the high-fitness mutants with a scoring function that employs an ensemble of methods including a PPLM.
    \item AFP-DE \cite{qin2023} uses a PPLM to sample variants and extract sequence embeddings. Iteratively trains a fitness predictor with the sampled variants and finetunes the PPLM with variants with high predicted fitness.
\end{itemize}
Furthermore, simulation of a single mutation walk (SMW) \cite{wu2019} is included in the comparison to serve as a baseline with no use of ML methods. Implementation details of the SMW simulation can be found in \cite{thesis}. Lastly, one conceptually different optimization MLDE method is included in the comparison. Neighborhood Search Directed Evolution (NSDE) \cite{thesis} performs a greedy graph search in a neighborhood graph constructed from the variants' sequence embeddings extracted by a PPLM.
\par
The regression-based methods are tested with a screening budget of 80 variants and two different splits between the part of the budget used for training and the rest of the budget left to screen variants with high predicted fitness.
SMW and the two optimization methods do not split the resources, so Table \ref{tbl:afpde_comparison} contains just a single result for these methods.
\begin{table}
    \centering
    \begin{tabular}{|c|c|c|}
        \hline
        \rowcolor[HTML]{dadada}
        Dataset & GB1 & PhoQ\\
        \hline
        \rowcolor[HTML]{dadada}
        Screening budget & $(24 + 56) \mid (48 + 32)$ & $(24 + 56) \mid (48 + 32)$\\
        \hline
        SMW & $3.90$ & $18.44$\\
        MLDE & $3.93 \mid 4.43$ & $\phantom{0}6.55 \mid 13.23$\\
        ftMLDE (EVmut.) & $4.99 \mid 5.27$ & $22.04 \mid \phantom{0}8.68$\\
        ftMLDE (trans.) & $4.98 \mid 5.31$ & $17.77 \mid 26.18$\\
        CLADE & $4.88 \mid 3.92$ & $21.51 \mid 25.65$\\
        CLADE 2.0 & $4.36 \mid 6.01$ & $24.45 \mid 21.51$\\
        AFP-DE & $6.20 \mid 6.20$ & $24.98 \mid 28.19$\\
        NSDE & $4.54$ & $20.71$\\
        BOES & $\textbf{7.28}$ & $\textbf{37.94}$\\
        \hline
    \end{tabular}
    \caption{Comparison of BOES with SOTA regression-based methods: maximum fitness obtained with 80 screened variants starting from the wild-type protein is reported. Regression-based methods split the screening budget between training and exploitation (24 + 56 or 48 + 32). Optimization methods screen all 80 variants during the optimization procedure.}
    \label{tbl:afpde_comparison}
\end{table}
\par
Table \ref{tbl:afpde_comparison} shows a clear dominance of the proposed BOES method. The results confirm that the optimization approach to MLDE can be more efficient than methods with a regression objective.

\subsection{Robustness to the Starting Protein}
While the performance on the wild-type protein corresponds to the practical use of MLDE algorithms, making conclusions about the methods' performance based on a single run, albeit on two different datasets, would be ill-advised. The results obtained from such a limited evaluation can be strongly skewed by the properties of the specific dataset. Especially local-search methods, like the SMW baseline or the NSDE method based on a KNN graph, could potentially show wildly different efficiency based on the relative position of the starting variant, the global optimum, and any local optima located between them in the sequence space.
\par
To ensure that the methods' hyperparameters are not over-fitted to the path from the wild-type variant to the global optimum, a test of robustness to the starting protein was conducted. This test also gives helpful insight into the variance in performance of the DE methods. Among the tested methods were the BOES method, the aforementioned NSDE method \cite{thesis}, a perceptron-training method based on the AFP-DE procedure \cite{qin2023} with a different \textit{exploration} stage implementation, and two simulations of classical DE methods without the use of ML: SMW and Recombination \cite{wu2019}. Implementation details of the tested methods are included in \cite{thesis}.
\par
\begin{figure*}
    \begin{subfigure}[t]{0.5\textwidth}
        \centering
        \includegraphics[width=\linewidth]{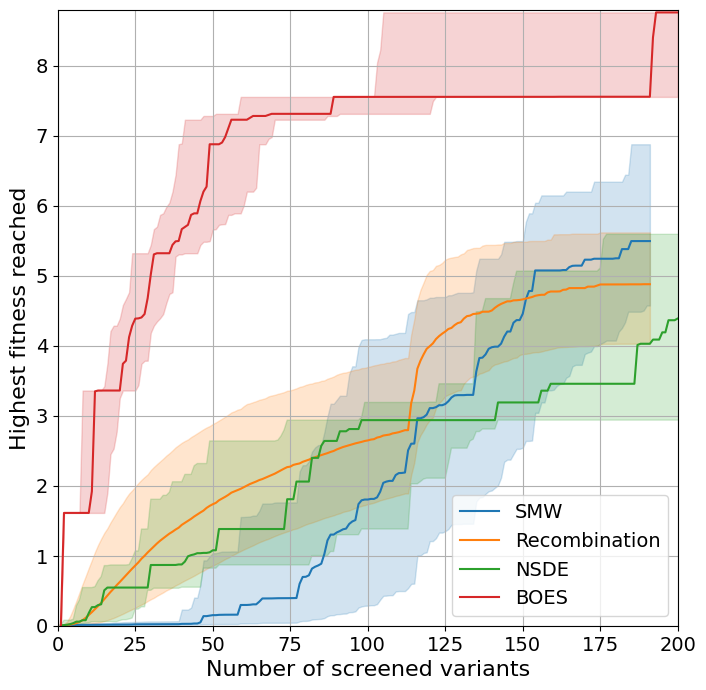}
        \captionsetup{justification=centering}
        \caption{Best-so-far fitness rogressions on GB1 dataset.}
        \label{fig:gb1_quartiles}
    \end{subfigure}
    \begin{subfigure}[t]{0.5\textwidth}
        \centering
        \includegraphics[width=\linewidth]{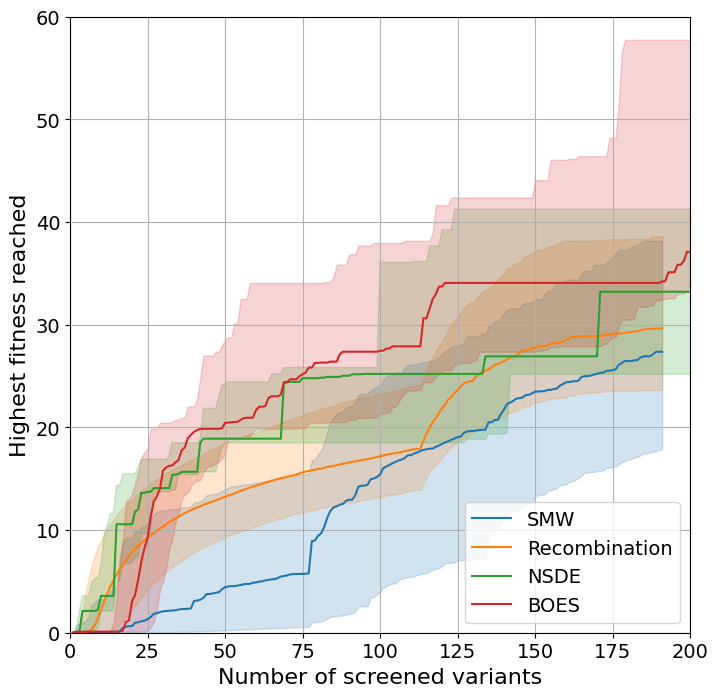}
        \captionsetup{justification=centering}
        \caption{Best-so-far fitness rogressions on PhoQ dataset.}
        \label{fig:phoq_quartiles}
    \end{subfigure}
    \vskip\baselineskip
    \begin{subfigure}[t]{0.5\textwidth}
        \centering
        \includegraphics[width=\linewidth]{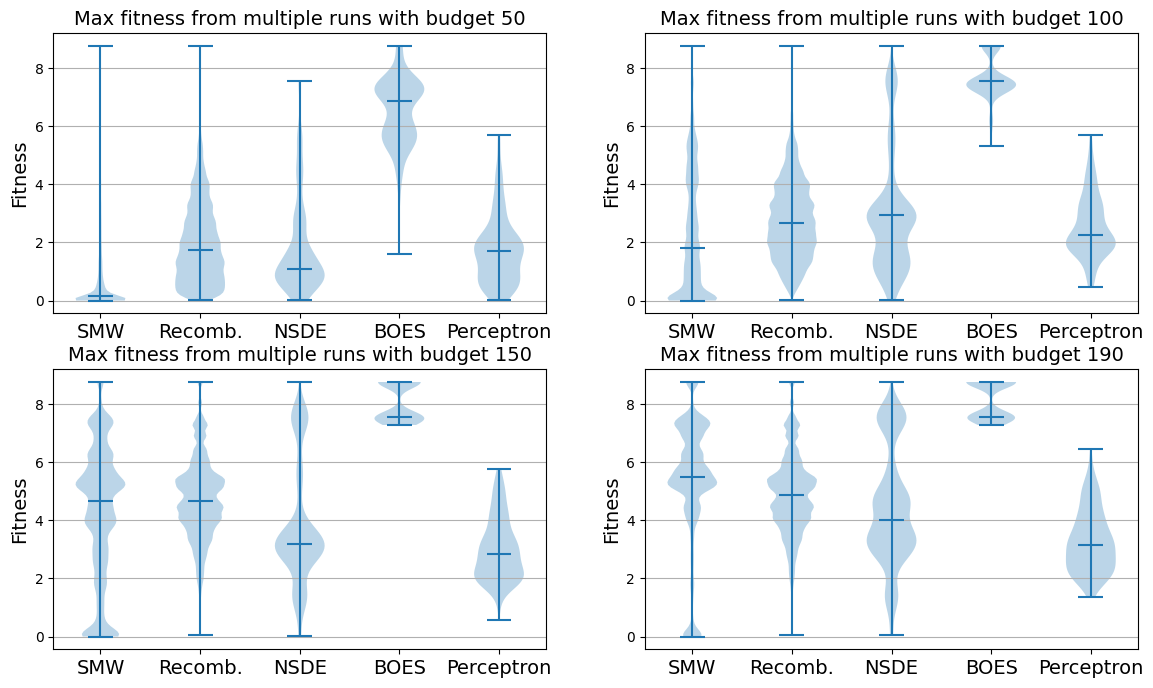}
        \captionsetup{justification=centering}
        \caption{Distributions of the highest achieved fitness on GB1 dataset.}
        \label{fig:violin_gb1}
    \end{subfigure}
    \begin{subfigure}[t]{0.5\textwidth}
        \centering
        \includegraphics[width=\linewidth]{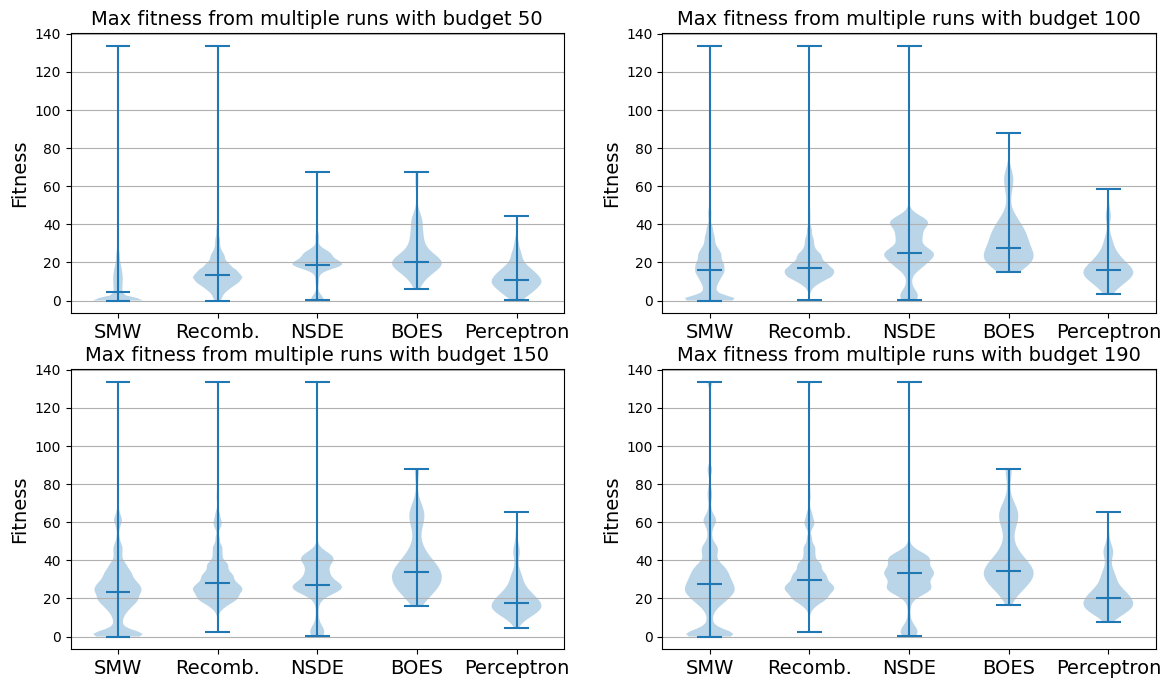}
        \captionsetup{justification=centering}
        \caption{Distributions of the highest achieved fitness on PhoQ dataset.}
        \label{fig:violin_phoq}
    \end{subfigure}
    \caption{Resulting fitness over multiple runs from sampled starting variants. (a, b) Best-so-far fitness progressions. Bold line represents median, highlighted areas correspond to interquartile range. (c, d) Distributions of the highest achieved fitness with different screening budgets.}
    \label{fig:de}
\end{figure*}

The evaluation was carried out by running each method repeatedly with a different starting variant for 200 to 160,000 runs, based on the computational demand of each method. The first quartile, median, and third quartile values of the highest achieved fitness by each method are reported in Fig. \ref{fig:gb1_quartiles} and Fig. \ref{fig:phoq_quartiles} for the GB1 and PhoQ dataset, respectively. Additionally, distributions of the highest achieved fitness at 50, 100, 150, and 190 screened variants are reported in Fig. \ref{fig:violin_gb1} and Fig. \ref{fig:violin_phoq}.
The starting variant is also counted towards the number of screened variants.

\par
The median fitness curve of the BOES method in Fig. \ref{fig:gb1_quartiles} shows that BOES usually finds the globally optimal variant in under 200 screened variants even when starting from a different, non-functional variant in the GB1 sequence space. The violin plots in Fig. \ref{fig:violin_gb1} and Fig. \ref{fig:violin_phoq} illustrate the superiority of our method further and show, that with increasing screening budget, BOES gives an increasing lower bound on the probable resulting fitness, whereas the other evaluated methods can strongly underperform more often.

\subsection{Ablation Study}
\textbf{Advantage of Embedding Space} To evaluate the effect of the informative input space on the performance of the BO procedure, we compare our results to a method proposed in \cite{frisby2020fold} which, in its most simple form, performs BO with a GP model and EI acquisition function directly in the protein sequence space. The distance between two variants is computed simply from one-hot encoding of the amino acids at mutated positions. This approach corresponds to the proposed BOES method in terms of the employed model and acquisition function but uses a straightforward definition of the input space and kernel function in place of the embedding space. This makes the method an ideal candidate for evaluation of the effect of the embedding space on the performance of BOES. The average results of the SMW baseline, BOES method, and the aforementioned
method with one-hot encoding, denoted as \textit{GP+EI}, are included in Table \ref{tbl:onehot_comparison}.
\begin{table}
    \centering
    \begin{tabular}{|c|c|c|}
        \hline
        \rowcolor[HTML]{dadada}
        \textbf{GB1 dataset} & Avg maximum fitness & Screening budget\\
        \hline
        SMW & $5.35 \pm 2.14$ & $191$\\
        GP+EI & $7.28 \phantom{\,\,\,\pm 0.00}$ & $20 + 191$\\
        BOES & $\textbf{8.14} \pm 0.62$ & $191$\\
        \hline
    \end{tabular}
    \caption{Comparison of BOES with BO conducted in the original protein sequence space: Mean of the maximum obtained fitness from multiple runs is reported. Results of the implemented methods are accompanied by standard deviation.}
    \label{tbl:onehot_comparison}
\end{table}
It is important to note that in the \textit{GP+EI} method from \cite{frisby2020fold}, the model is initially trained on 20 randomly selected variants before the first iteration of BO, which are not counted towards the screening budget, skewing the comparison in its favor. One last note-worthy distinction between BOES and the method reported in \cite{frisby2020fold} is that, unlike BOES, this method selects new variants for screening in batches of 19.
Comparison in Table \ref{tbl:onehot_comparison} decidedly confirms a positive effect of employing BO in the embedding space over the original sequence space with a one-hot encoding-based kernel function.

\par
\textbf{Other Informative Input Spaces}
The conducted comparison to a BO-based method defined on the original sequence space proves the positive effect of the innovative input space qualitatively. However, a comparison to another state-of-the-art BO-based method with a different, yet also informative, input space can help assess the performance of the proposed method quantitatively. The ODBO framework \cite{cheng2022odbo} employs BO for DE in combination with a novel encoding of amino acids based on the fitness of observed variants with the specific amino acids at the specified mutation position.
In Table \ref{tbl:encode_comparison}, results of the BOES method with a screening budget of 50 variants are compared to a classical BO procedure with a GP model and the positional amino-acid encoding of variants from \cite{cheng2022odbo} and to a trust region BO procedure (TuRBO) \cite{eriksson2019scalable} with the same model and encoding.
\begin{table}
  \centering
    \begin{tabular}{|c|c|c|}
        \hline
        \rowcolor[HTML]{dadada}
        \textbf{GB1 dataset} & Avg maximum fitness & Screening budget\\
        \hline
        NaiveBO + GP & $6.40 \pm 0.79$ & $40 + 50$\\
        TuRBO & $\textbf{6.57} + 1.02$ & $40 + 50$\\
        BOES & $6.47 \pm 1.15$ & $\textbf{50}$\\
        \hline
    \end{tabular}
  \caption{Comparison of BOES to BO conducted with a different informative sequence representation: Mean of the maximum obtained fitness from multiple runs is reported with the standard deviation.}
    \label{tbl:encode_comparison}
\end{table}
\par
A critical difference between the two methods of sequence-space representation is that the positional amino-acid encoding proposed in \cite{cheng2022odbo} requires an initial dataset of screened variants in which each amino acid appears at each mutation site at least two times, while the PPLM-extracted embedding space used in BOES requires no screened variants for its construction. The original paper presents a solution to this obstacle in the form of an initial sampling algorithm, which for the GB1 dataset constructs an initial set of 40 variants. This means that while each of the BO procedures compared in Table \ref{tbl:encode_comparison} is provided with a screening budget of 50 variants, the construction of the encoding that precedes the two procedures from \cite{cheng2022odbo} requires an additional 40 screening experiments, which the BOES method saves in comparison.
\par
Results in Table \ref{tbl:encode_comparison} reveal that all three compared BO-based methods produce proteins of extremely similar quality with BOES outperforming the other classical BO method, labeled NaiveBO, and the trust region variant slightly outperforming BOES. It should be noted that a trust region variant of BOES could also be implemented, which would most probably improve the original BOES method's performance. Similarly, the authors of the compared BO method \cite{cheng2022odbo} propose two additional improvements: prescreening outlier detection via XGBOD \cite{zhao2018xgbod} and employing a BO procedure robust to outliers \cite{martinez2018practical}. The variant of the authors' method with these improvements outperforms BOES, but the improvements could also be combined with BOES. Adding the prescreening outlier detection step requires a set of already screened variants. To circumvent this, the outlier detection could be enabled after a certain number of BOES iterations. Additionally, the XGBOD method could be replaced with an unsupervised outlier detection method in the initial iterations of BOES. A version of BOES with the additional improvements from \cite{cheng2022odbo} can be expected to yield similar results to the full version of ODBO \cite{cheng2022odbo} while saving screening costs on the construction of sequence representation.

\subsection{Visualizing the Embedding Space}
The BOES method operates on a PPLM-extracted sequence embedding space instead of using the raw sequences of amino acids. It is crucial that the embedding space provides a sensible metric of similarity between variants as well as encodes useful information about the variants' properties. To ensure that this assumption holds, the embedding space was visualized with dimensionality reduction methods.
\par
First, a joint principal component analysis (PCA) was conducted on sequence embeddings of all variants from both datasets (GB1 and PhoQ) as a sanity check. The PCA confirmed that the two datasets are easily separable. The first principal component alone accounts for 97.4~\% of variance in the joint distribution and separates the two datasets into two clear clusters.
\par
Next, PCA was conducted for each dataset separately to visually confirm whether expected features of the sequence space, like local maxima and distinguishable areas with low/high fitness, are present in the embedding space. Results of PCA in both of the datasets revealed that the first two principal components together explain roughly 40~\% of the variance. That is a very large portion, considering that the ESM-1b embedding space has 1280 dimensions. The PCA analyses of the standalone datasets both showed one large area with functional variants.
\begin{figure*}
    \def\imgScale{0.55}
    \begin{subfigure}[t]{0.5\textwidth}
        \centering
        \includegraphics[scale=\imgScale]{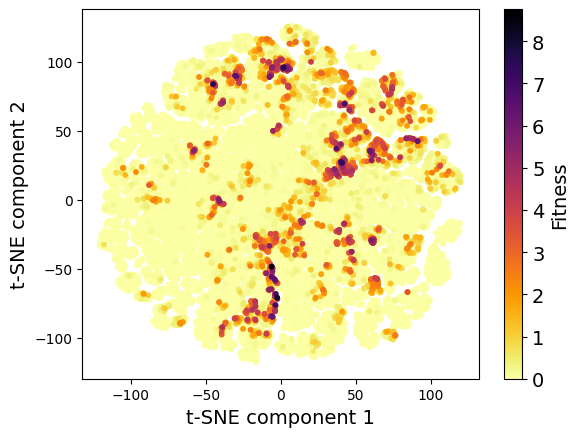}
        \captionsetup{justification=centering}
        \caption{t-SNE of GB1 embedding space with true fitness.}
        \label{fig:gb1_tsne}
    \end{subfigure}
    \begin{subfigure}[t]{0.5\textwidth}
        \centering
        \includegraphics[scale=\imgScale]{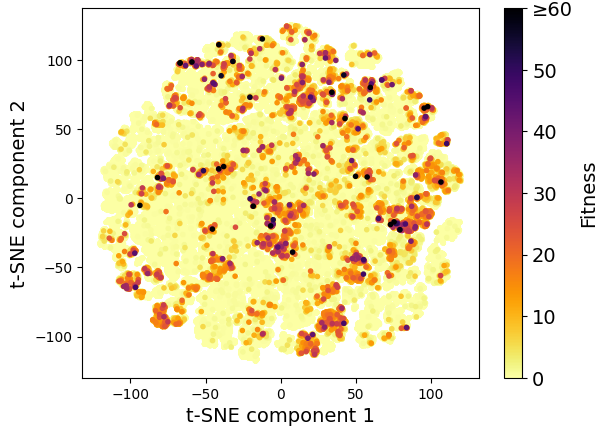}
        \captionsetup{justification=centering}
        \caption{t-SNE of PhoQ embedding space with true fitness.}
        \label{fig:phoq_tsne}
    \end{subfigure}
    \vskip\baselineskip
    \begin{subfigure}[t]{0.5\textwidth}
        \centering
        \includegraphics[scale=\imgScale]{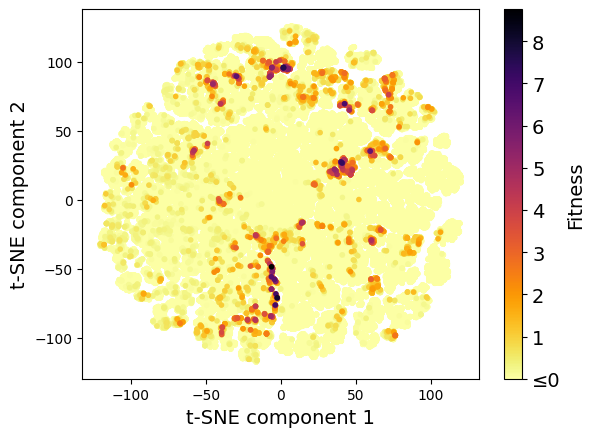}
        \captionsetup{justification=centering}
        \caption{t-SNE of GB1 embedding space with predicted fitness.}
        \label{fig:gb1_tsne_boes}
    \end{subfigure}
    \begin{subfigure}[t]{0.5\textwidth}
        \centering
        \includegraphics[scale=\imgScale]{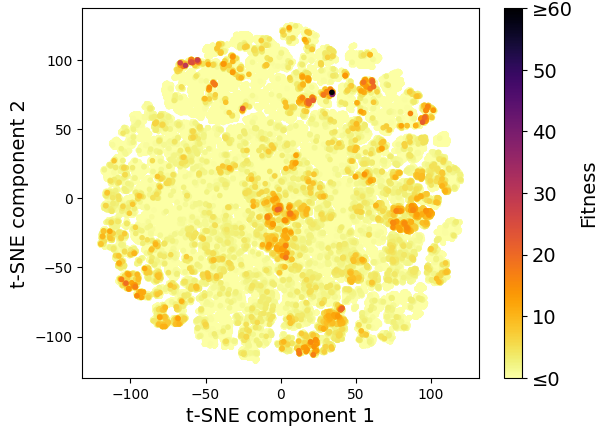}
        \captionsetup{justification=centering}
        \caption{t-SNE of PhoQ embedding space with predicted fitness.}
        \label{fig:phoq_tsne_boes}
    \end{subfigure}
    \caption{Visualisation of (a, c) GB1 and (b, d) PhoQ embedding space extracted with ESM-1b PPLM with (a, b) true fitness and (c, d) fitness predicted by GP model trained on 384 screened variants in BOES run from the wild type protein.}
    \label{fig:dimred}
\end{figure*}
To assess whether the embedding space is capable of capturing local maxima in fitness landscapes, the embedding space of each dataset was visualized with the t-SNE method, which emphasizes maintaining low distances between close data points, preserving local clusters. The t-SNE visualization is plotted in Fig. \ref{fig:gb1_tsne} for the GB1 dataset and in Fig. \ref{fig:phoq_tsne} for the PhoQ dataset. Both figures confirm the presence of local clusters of high-fitness variants.

\subsection{Modelling the Embedding Space}
As a secondary result, the fitness landscape modeled by the BOES method was visualized alongside the initial t-SNE plots in Fig. \ref{fig:dimred}. Fig. \ref{fig:gb1_tsne_boes} and  Fig. \ref{fig:phoq_tsne_boes} show fitness predicted by a GP model trained on 384 screened variants by the BOES method when initiated with the wild-type protein of the GB1 and PhoQ datasets, respectively.
The visualization of predicted fitness shows that BOES was able to identify multiple local clusters of high-fitness variants in both datasets. Especially the result on the GB1 dataset in Fig. \ref{fig:gb1_tsne_boes} reveals that almost all of the major clusters were identified by BOES. High normalized discounted cumulative gain (NDCG) values (GB1: 0.88, PhoQ: 0.79) confirm that BOES effectively models the fitness landscape to rank high-fitness variants, which is essential in MLDE.

\section{Conclusion}
In this paper, we have presented a novel method of machine-learning-assisted directed evolution (MLDE), termed Bayesian optimization in embedding space (BOES). Feasibility of the proposed method was confirmed in silico on two datasets. Our method outperforms SOTA MLDE methods with a regression objective. Moreover, the informative representation of the input space based on the sequence embeddings extracted by a pre-trained protein language model (PPLM) has been shown to significantly improve the performance of Bayesian optimization (BO) over optimization in the original protein sequence space. The BOES method produces proteins of comparable quality to other state-of-the-art BO-based methods that employ different informative representations of the input space while significantly reducing screening costs since there is no screening needed within the construction of the sequence representation. This improvement can result in saved resources on experimental costs or more resources for additional iterations of DE, yielding better results with the same amount of conducted screening in total. For future development, we suggest combining the innovative input space representation proposed in this paper with the improvements to the standard BO procedure suggested in \cite{cheng2022odbo}. We order the suggestions based on their effect on the performance of the ODBO method \cite{cheng2022odbo}. Firstly, conducting prescreening outlier detection via \textit{Extreme Gradient Boosting Outlier Detection} \cite{zhao2018xgbod} in later iterations. Secondly, implementing a BO procedure robust to outliers based on \cite{martinez2018practical} and finally, employing trust region BO \cite{eriksson2019scalable}.

\section*{Acknowledgment}
Special thanks go to Šimon Soldát for useful discussions and insight into Bayesian optimization.

This research work was supported by the Grant Agency of the Czech Technical University in Prague, grant No. SGS23/184/OHK3/3T/13.
The access to the computational infrastructure of the OP VVV funded project\newline
CZ.02.1.01/0.0/0.0/16\_019/0000765 ``Research Center for Informatics'' is also gratefully acknowledged.

\bibliographystyle{IEEEtran}
\bibliography{IEEEabrv,./references}

\end{document}